\documentclass[times, 10pt,twocolumn]{article}
\usepackage{latex8}
\usepackage{times}
\usepackage{balance}
\usepackage{amsmath,epsfig}
\usepackage{rotating}
\usepackage{amssymb}
\usepackage{graphicx,amsfonts}
\usepackage[T1]{fontenc}
\usepackage{amsthm}
\usepackage{url}
\usepackage{multirow}

\newcommand{\cH}{\mathcal{H}}
\newcommand{\cX}{\mathcal{X}}

\newcommand{\bx}{\boldsymbol{x}}
\newcommand{\bh}{\boldsymbol{h}}
\newcommand{\by}{\boldsymbol{y}}

\newcommand{\beq}{\begin{equation}}
\newcommand{\eeq}{\end{equation}}
\newcommand{\beqn}{\begin{eqnarray}}
\newcommand{\eeqn}{\end{eqnarray}}
\newcommand{\beqns}{\begin{eqnarray*}}
\newcommand{\eeqns}{\end{eqnarray*}}

\newcommand{\R}{\mathbb{R}}

\makeatletter
\newcommand{\bdiv}{\mathop{\operator@font div}}
\makeatother

\makeatletter
\newcommand{\diag}{\mathop{\operator@font diag}}
\makeatother

\makeatletter
\newcommand{\conv}{\mathop{\operator@font conv}}
\makeatother

\makeatletter
\newcommand{\sign}{\mathop{\operator@font sign}}
\makeatother \makeatletter

\makeatletter
\newcommand{\proj}{\mathop{\operator@font proj}}
\makeatother \makeatletter

\makeatletter
\newcommand{\spa}{\mathop{\operator@font span}}
\makeatother \makeatletter

\makeatletter
\newcommand{\epi}{\mathop{\operator@font epi}}
\makeatother \makeatletter

\makeatletter
\newcommand{\dom}{\mathop{\operator@font dom}}
\makeatother \makeatletter

\newtheorem{thm}{Theorem}[section]

\theoremstyle{remark}

\theoremstyle{definition}

\theoremstyle{definition}

\begin{document}

\title{Edge Preserving Image Denoising in Reproducing Kernel Hilbert Spaces}

\author{Pantelis Bouboulis, Sergios Theodoridis\\
\emph{Department of Informatics and Telecommunications,}\\
\emph{University of Athens,}\\
\emph{Athens, Greece.}\\
\emph{\{bouboulis,stheodor\}@di.uoa.gr}\\
\and
Konstantinos Slavakis\\
\emph{Department of Telecommunications}\\
\emph{Science and Technology,}\\
\emph{University of Peloponnese,}\\
\emph{Tripolis, Greece.}\\
\emph{slavakis@uop.gr}
}

\maketitle
\thispagestyle{empty}

\begin{abstract}
The goal of this paper is the development of a novel approach for the problem of Noise Removal, based on the theory of Reproducing Kernels Hilbert Spaces (RKHS). The problem is cast as an optimization task in a RKHS, by taking advantage of the celebrated semiparametric Representer Theorem. Examples verify that in the presence of gaussian noise the proposed method performs relatively well compared to wavelet based technics and outperforms them significantly in the presence of impulse or mixed noise.
\end{abstract}

\section{Introduction}\label{SEC:Intro}
The problem of noise removal from a digitized image is one of the most fundamental ones in digital image processing. So far, various techniques have been proposed to deal with it. Among the most important methodologies are, for example, the Wavelet-based image denoising methods, which dominates the research in recent years \cite{Poly, DaFoKatEg}. In this paper we propose a novel approach which (to our knowledge) has not been considered before. We employ the well known powerful tool of kernels.

In kernel methodology the notion of the Reproducing Kernel Hilbert Space (RKHS) plays a crucial role. A RKHS, is a rich construct (roughly, a smooth space with an inner product), which has been proven to be a very powerful tool for non linear processing \cite{SchoSmo, Theokou}. In the denoising problem, we exploit a useful property of RKHS, the \textit{representer theorem} \cite{SchoSmo}. It states that the minimizer of any optimization task in $\cH$, with a cost function of a certain type, has a finite representation in $\cH$. We recast the image denoising problem as an optimization task of this type and use the semi-parametric version of the representer theorem. The latter, allows for explicit modeling of the edges in an image. In such a way we can deal with the smoothness which is, implicitly, imposed by the "smooth" nature of RKHS.

Though there has been some work exploring the use of kernels in the denoising problem, the methodology presented here is fundamentally different. In \cite{TaFaMil}, the notion of kernel regression has been adopted. The original image is formulated as a Taylor approximation series around a center, $\bx_i$, and data adaptive kernels are used, as weighted factors, to penalize distances away from $\bx_i$. In a relatively similar context, kernels have been employed by other well known denoising methods (such as \cite{BuaCoMor}). Kernels were also used in the context of RKHS in \cite{Dalli, KimFraScho}. However, the obtained results were not satisfying, especially around edges. It is exactly this drawback that is addressed by our method.

\section{Mathematical Preliminaries}
We start with some basic definitions regarding RKHS. Let $X$ be a non empty set with $\bx_1,\dots,\bx_N \in X$.
Consider a Hilbert space $\cH$ of real valued functions $f$ defined on a set $X$, with a corresponding inner product $\langle\cdot,\cdot\rangle_\cH$. We will call $\cH$ as a \textit{Reproducing Kernel Hilbert Space} - RKHS, if there exists a function, known as kernel, $\kappa:X\times X\rightarrow\R$ with the following two properties:
\begin{enumerate}
\item For every $\bx\in X$, $\kappa(\bx,\cdot)$ belongs to $\cH$.
\item $\kappa$ has the so called \textit{reproducing property}, i.e.
$f(\bx)=\langle f,\kappa(\bx,\cdot)\rangle_\cH, \textrm{ for all } f\in\cH$.
In particular $\kappa(\bx,\by)=\langle \kappa(\bx,\cdot),\kappa(\by,\cdot)\rangle_\cH$.
\end{enumerate}

In can been shown that the kernel $\kappa$ produces the entire space $\cH$, i.e. $\cH=\overline{\spa\{\kappa(\bx,\cdot)|\bx\in X\}}.$
There are several kernels that are used in practice (see \cite{SchoSmo}). In this work, we focus on one of the most widely used, the Gaussian Kernel: $$\kappa(\bx,\by)=exp\left(-\frac{\|\bx-\by\|^2}{2\sigma^2}\right), \sigma>0,$$
due to some additional properties that it admits.

One of the many powerful tools in kernel theory is the application of the semi-parametric representer theorem to \textit{regularized risk minimization} problems (see \cite{SchoSmo}):
\begin{thm}\label{THE:semi_repr}
Denote by $\Omega_1,\Omega_2:[0,\infty)\rightarrow\R$, two strictly monotonic increasing functions, by $\cX$ a set and by $c:(\cX\times\R^2)^m\rightarrow\R\cup\{\infty\}$
an arbitrary loss function. Furthermore, consider a set of $M$ real-valued functions $\{\psi_k\}_{k=1}^M:\cX\rightarrow\R$, with the property that the $N\times M$ matrix $(\psi_p(\bx_n))_{n,p}$ has rank $M$. Then any $\tilde f:=f+h$, with $f\in\cH$ and
$h\in \mathfrak{H}=\textrm{span}\{\psi_k\},$
minimizing the regularized risk functional
\begin{align*}
c\left((\bx_1,z_1,f(\bx_1)),\dots,(\bx_N,z_N,f(\bx_N)\right)\\+\phantom{2}\Omega_1\left(\|f\|_\cH\right) + \Omega_2\left(\|h\|_{\mathfrak{H}}\right)
\end{align*}
admits a representation of the form
\begin{align}\label{EQ:semi_repr_2}
\tilde f(\bx)=\sum_{n=1}^N \alpha_n \kappa(\bx_n,\bx) + \sum_{k=1}^M\beta_k\psi_k(\bx).
\end{align}
\end{thm}

Usually the regularization term $\Omega(f)$ takes the form $\Omega(f)=\frac{1}{2}\|f\|_\cH^2$. In the case of the RKHS produced by the gaussian Kernel we can prove that
\begin{align}\label{EQ:regul}
\|f\|_\cH=\int_\cX\sum_n\frac{\sigma^{2n}}{n!2^n}(O^n f(\bx))^2d\bx,
\end{align}
with $O^{2n}=\Delta^n$ and $O^{2n+1}=\nabla\Delta^n$, $\Delta$ being the Laplacian and $\nabla$ the gradient operator (see \cite{SchoSmo}). Thus, we see that the regularization term "penalizes" the derivatives of the minimizer. This results to a very smooth solution of the regularized risk minimization problem.

Note that according to theorem \ref{THE:semi_repr} the model of a function has two parts, one lying in the smooth RKHS space and another part $h$ which gives rise to the second term in the expansion (\ref{EQ:semi_repr_2}). It is exactly this term that is exploited by our method in order to explicitly model edges.

\begin{table*}[t]
\centering
\caption{Results on Boat corrupted by impulse noise.}\label{TAB:impulse}
\tiny
\begin{tabular}{|c|c|c|c|c|c|c|c|c|c|}
\hline
\textbf{Image}   &   \textbf{Noise}   & \textbf{noisy PSNR}    &   \textbf{Kernel Denoising}  &  \textbf{BiShrink} \cite{Poly}  & \textbf{K-SVD} \cite{ElAh06} & \textbf{SKR}  $L_1$\cite{TaFaMil} & \textbf{SKR} \cite{TaFaMil} & \textbf{BM3D} \cite{DaFoKatEg}\\\hline
  \multirow{4}{*} {\textbf{Boat}}
  &  20\%   & 18.56 dB  &  32.36 dB  & 22.59 dB & 26.46 dB & 31.85 dB & 28.35 dB & 29.45 dB\\
  &  30\%   & 16.77 dB  &  30.66 dB  & 25.07 dB & 26.79 dB & 30.85 dB & 27.05 dB & 28.29 dB\\
  &  40\%   & 15.52 dB  &  29.14 dB  & 25.40 dB & 26.08 dB & 29.51 dB & 25.85 dB & 27.26 dB\\
  &  50\%   & 14.55 dB  &  28.10 dB  & 25.09 dB & 25.38 dB & 27.73 dB & 24.90 dB & 26.61 dB\\\hline
\end{tabular}
\end{table*}

\begin{table*}[t]
\centering
\caption{Results on Lena corrupted by gaussian noise.}\label{TAB:gaussian}
\tiny
\begin{tabular}{|c|c|c|c|c|c|c|c|c|c|c|c|}
\hline
\textbf{Image}   &   \textbf{Noise}   & \textbf{noisy PSNR}    &   \textbf{Kernel Denoising}  &  \textbf{BiShrink} \cite{Poly}  & \textbf{BLS-GSM} \cite{Portilla2003} & \textbf{K-SVD} \cite{ElAh06} & \textbf{SKR} $L_1$ \cite{TaFaMil} & \textbf{SKR} \cite{TaFaMil} & \textbf{BM3D} \cite{DaFoKatEg}\\\hline
\multirow{3}{*} {\textbf{Lena}}
  &  $s=10$   & 28.12 dB  &  33.98 dB  & 34.33 dB  &  35.60 dB & 35.47 dB & 32.66 dB & 35.32 dB & 35.93 dB\\
  &  $s=20$   & 22.14 dB  &  31.12 dB  & 31.17 dB  &  32.65 dB & 32.36 dB & 29.23 dB & 32.62 dB & 33.00 dB\\
  &  $s=30$   & 18.72 dB  &  29.11 dB  & 29.35 dB  &  30.50 dB & 30.30 dB & 26.60 dB & 30.71 dB & 31.21 dB\\\hline
  \end{tabular}
\end{table*}

\section{Application to the denoising problem}\label{SEC:RKHS}

\begin{figure}
\begin{center}
\includegraphics[scale=0.3]{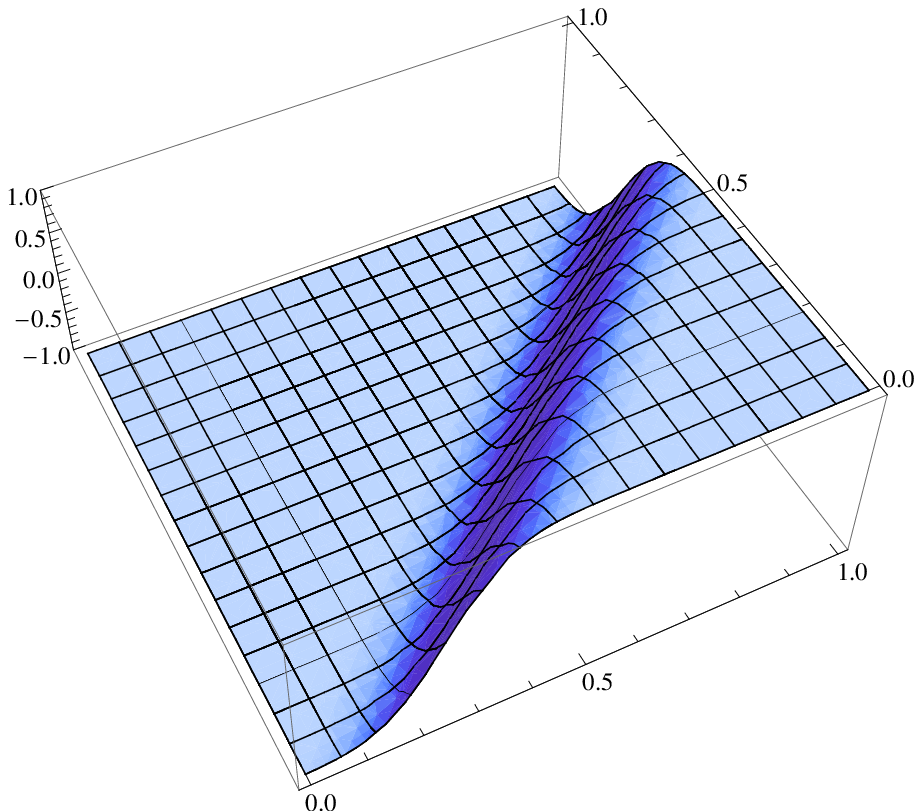}
\includegraphics[scale=0.3]{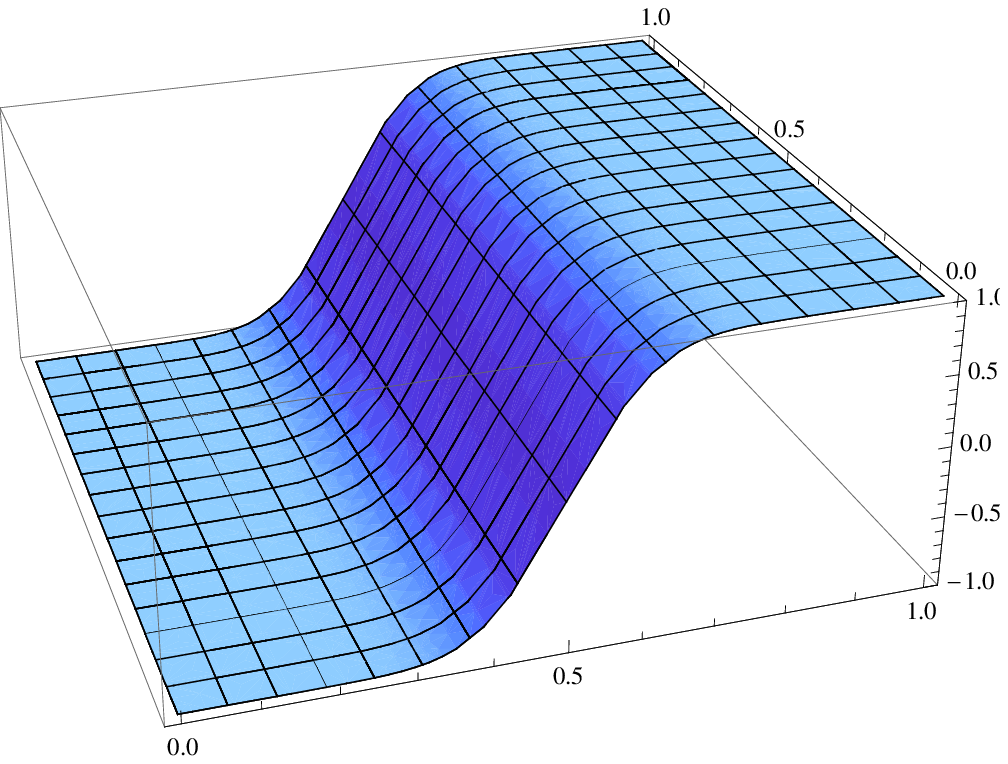}
\end{center}
\caption{Two of the functions $\psi_k$ that are used to represent edges.}\label{FIG:erfs}
\end{figure}

Let $f$ be the original image and $\hat f$ the noisy one (we consider them as continuous functions). Also, let $f_{i,j}$ and $\hat f_{i,j}$ be the restrictions of $f$ and $\hat f$ on the $N\times N$ orthogonal region centered at the pixel $(i,j)$ of each image accordingly ($N$ is an odd number). Our task is to find $f_{i,j}$ from the given samples of $\hat f_{i,j}$. For simplicity, we drop the $i,j$ indices and consider $f_{i,j}$ and $\hat f_{i,j}$ (which from now on will be written as $f$ and $\hat f$) as functions defined on $[0,1]^2$ (and zero elsewhere). The pixel values of the digitized image are given by $f(x_n,y_m)$ and $\hat f(x_n,y_m)$ where $x_n=n/(N-1)$, $y_m=m/(N-1)$ for $n,m=0,1,\dots,N-1$.

We consider a set of real valued functions $\{\psi_k,\; k=1,\dots,K\}$ with two variables suitable to represent edges; i.e., bivariate polynomials (which are controlled by the coefficients $h_0, h_1, h_2, h_3$) and functions of the form $\textrm{Erf}(a\cdot x+b\cdot y + c)$, where $\textrm{Erf}$ is the error function,
$$\textrm{Erf}(x)=\frac{2}{\sqrt{\pi}}\int_0^x e^{-t^2}dt,$$
for several suitable choices of $a$, $b$ and $c$ (see figure \ref{FIG:erfs}). Thus we formulate the regularized risk minimization problem as follows:
\begin{align}
&\mathop{\textrm{minimize}}_{f\in\cH,\;\boldsymbol{\beta}\in\R^K,\;\bh\in\R^4}\phantom{5}  \sum_{n=0}^{N-1}\sum_{m=0}^{M-1} \Big|f(x_n,y_m) +h_0 + h_1x_n + \nonumber\\
&+ h_2y_m + h_3x_ny_m + \sum_{k=1}^K \beta_k\psi_k(x_n,y_m) - \hat f(x_n,y_m)\Big| \phantom{2}  \nonumber\\
&+ \phantom{2}\frac{\lambda}{2}\|f\|^2_\cH \phantom{2} +\phantom{2}\frac{\mu}{2} \sum_{k=1}^K |\beta_k|^2 + \frac{\mu_1}{2}\sum_{l=1}^3 h_l^2.
\end{align}
Taking a closer look at the term $\frac{\lambda}{2}\|f\|^2_\cH$ according to equation (\ref{EQ:regul}), one sees that we actually penalize the derivatives of $f$ in a more influential fashion than the \textit{total variation} scheme, which is often used in wavelet-based denoising and penalizes only the first order derivatives. It turns out that in our method the use of the $L_1$ norm in the cost function, in combination with regularization, results in sparse modeling with respect to the $\beta$ coefficients. It should be noted that the use of the $L_1$ norm, also, in the regularization term leads to similar results.

The semi-parametric theorem \ref{THE:semi_repr} ensures that the minimizer will have a finite representation of the form:
\begin{align*}
\tilde f(x,&y)=\sum_{n=0}^{N-1}\sum_{m=0}^{M-1} \alpha_{n,m} \kappa((x_n,y_m),(x,y)) + \\
& + \sum_{k=1}^M\beta_k\psi_k(x,y) + h_0 + h_1x + h_2y + h_3xy.
\end{align*}
We can solve this problem using Polyak's Projected Subgradient Method \cite{Polyak}. We fix the regularization parameter $\lambda$ and adjust $\mu$ and $\mu_1$ so that they take small values around edges and large values in smooth areas.
In particular, as the algorithm moves from one pixel to the next, it decides whether the corresponding pixel centered region contains edges or not using the mean gradient of the specific region and then, it solves the corresponding minimization problem.

\section{Experimental Results}
\begin{sidewaysfigure*}[p]
\centering
\scalebox{0.25}
{\includegraphics{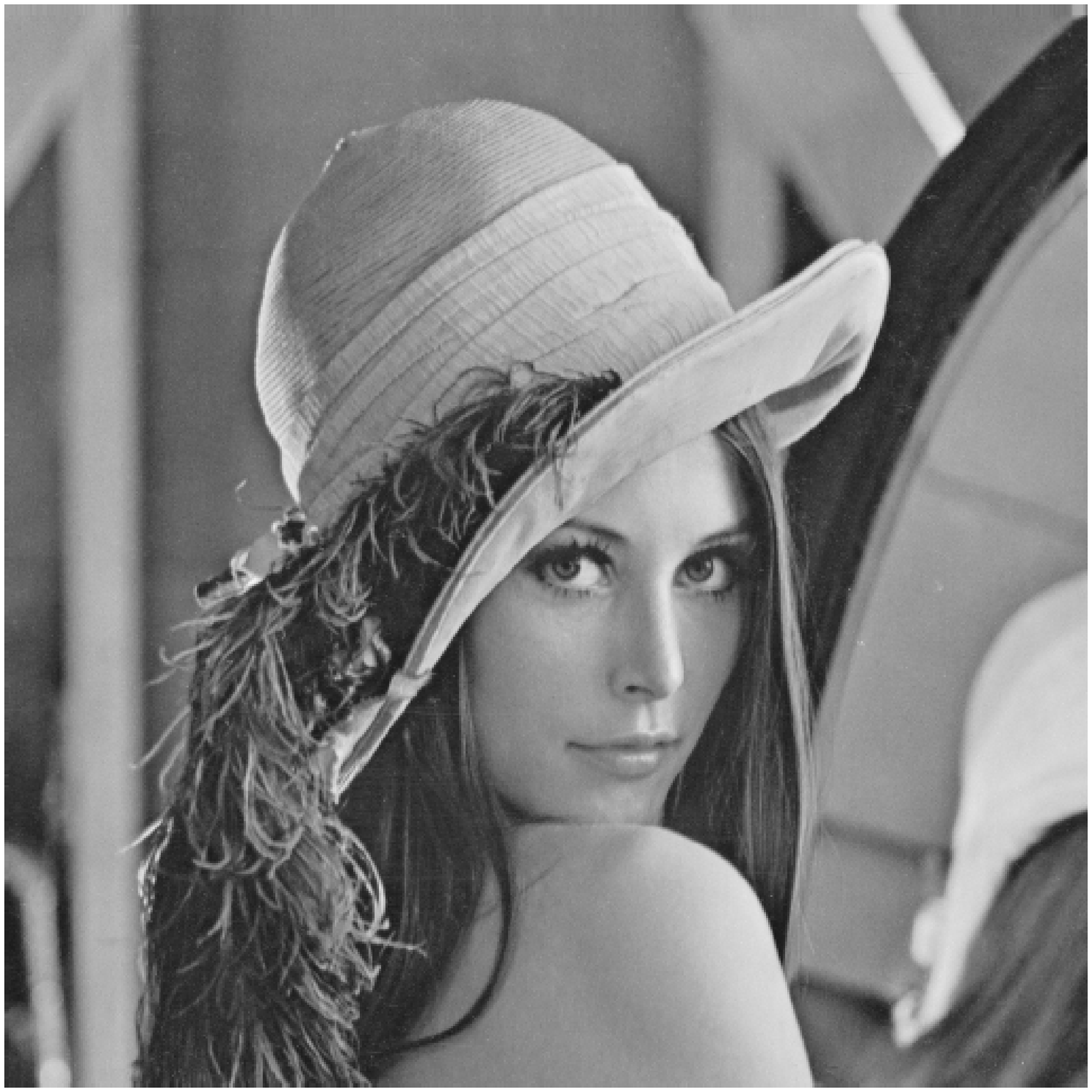}\phantom{2} \includegraphics{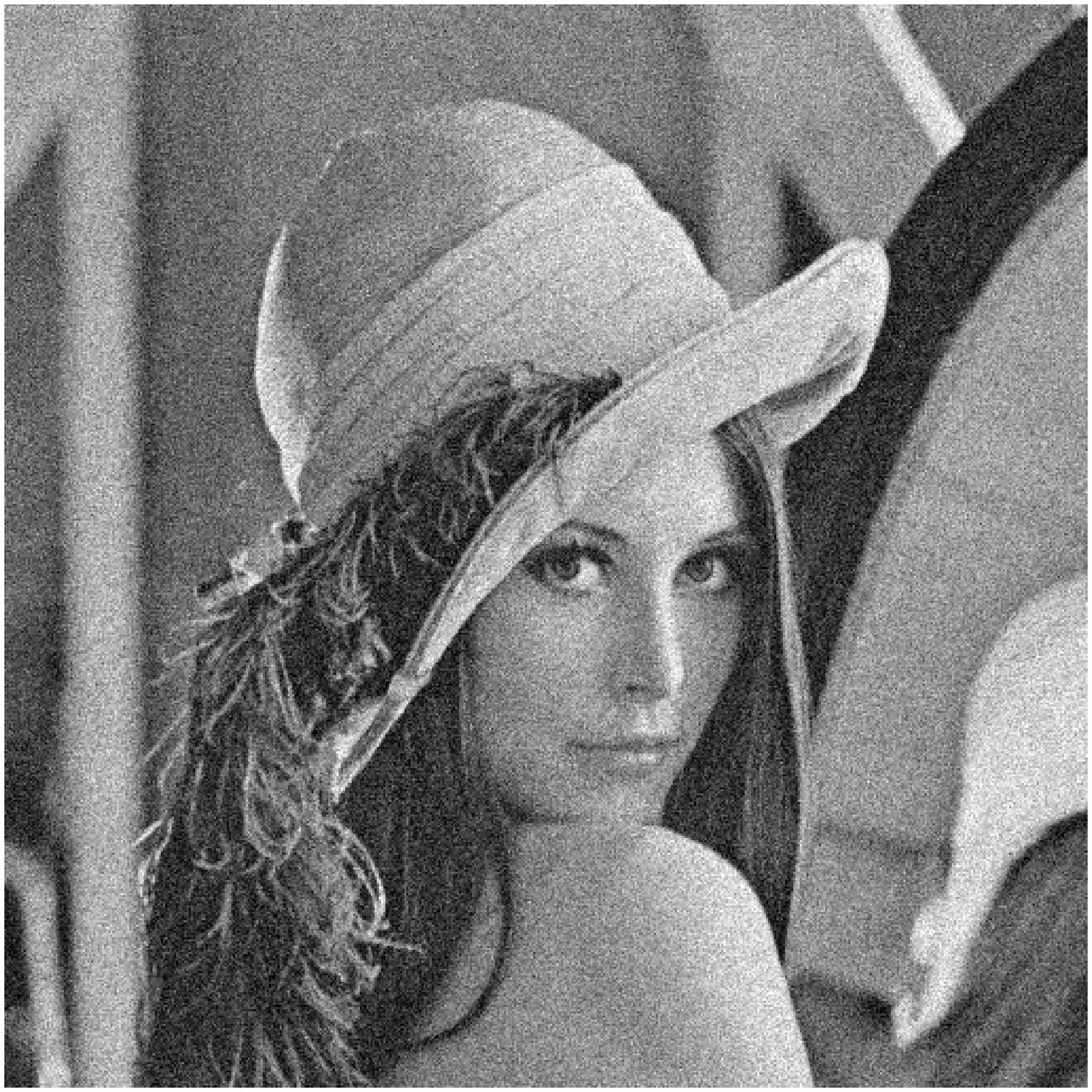}\phantom{2}
\includegraphics{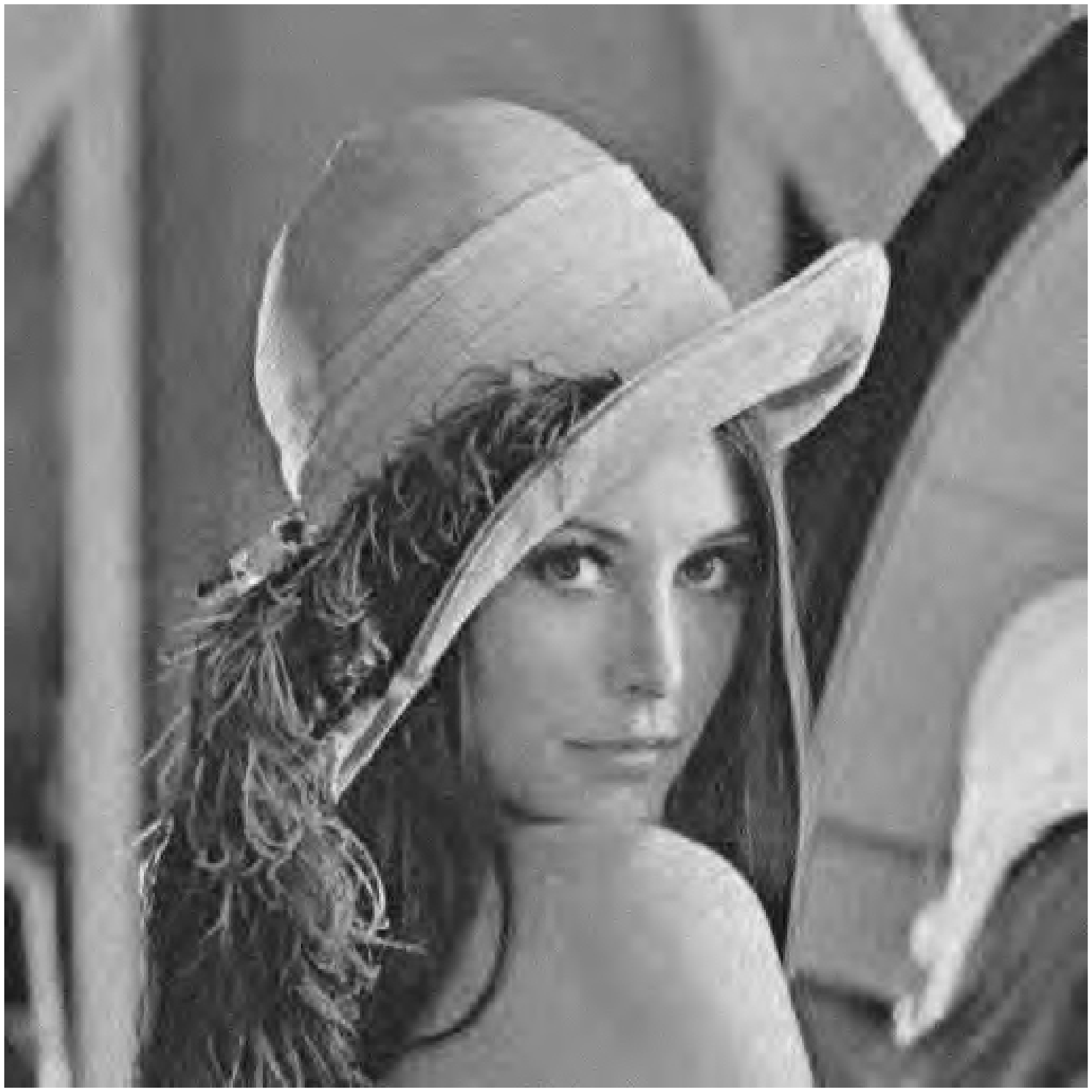}\phantom{2}
\includegraphics{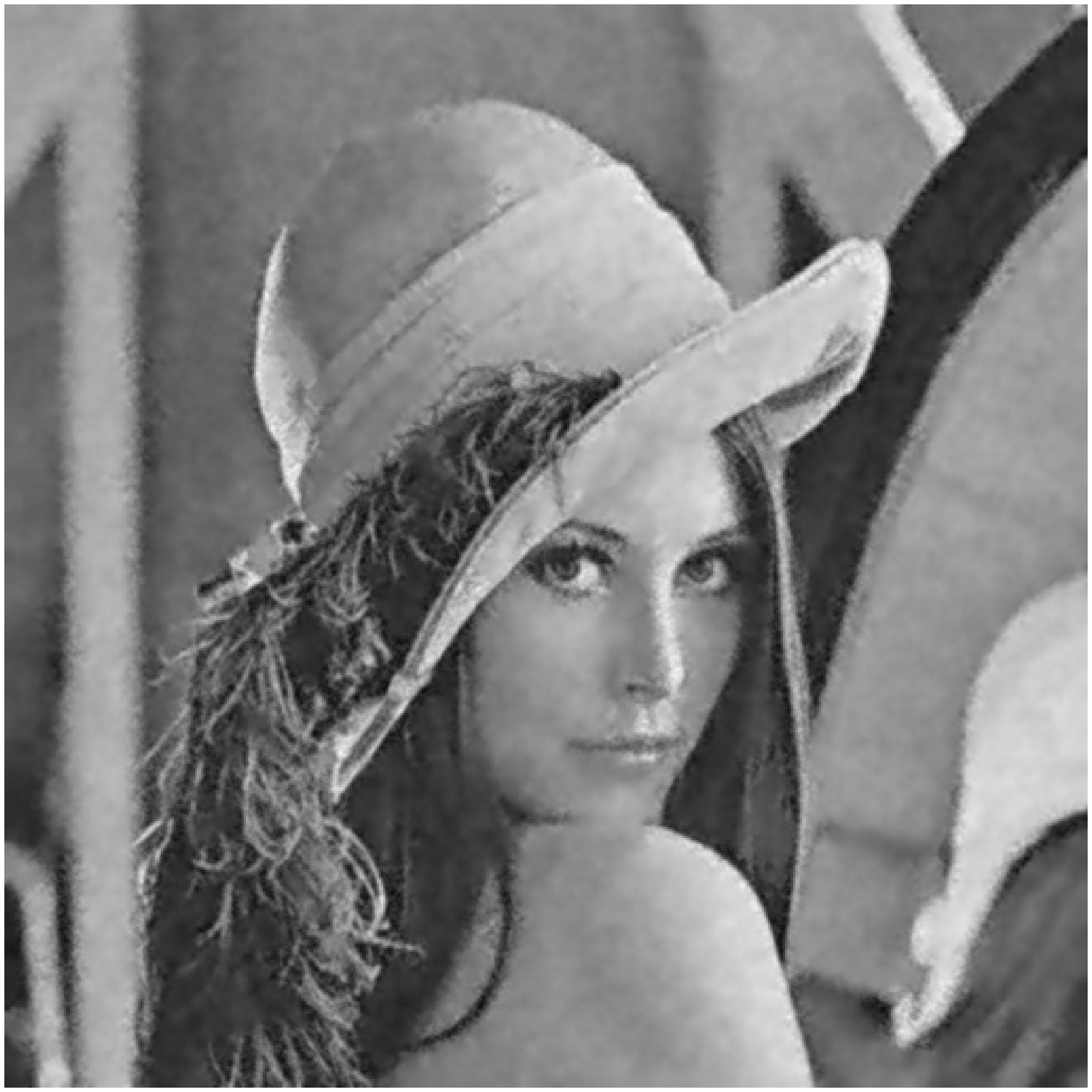}}
\centerline{(a) \hspace{13em}(b) \hspace{13em} (c) \hspace{13em} (d)}

\scalebox{0.25}
{\includegraphics{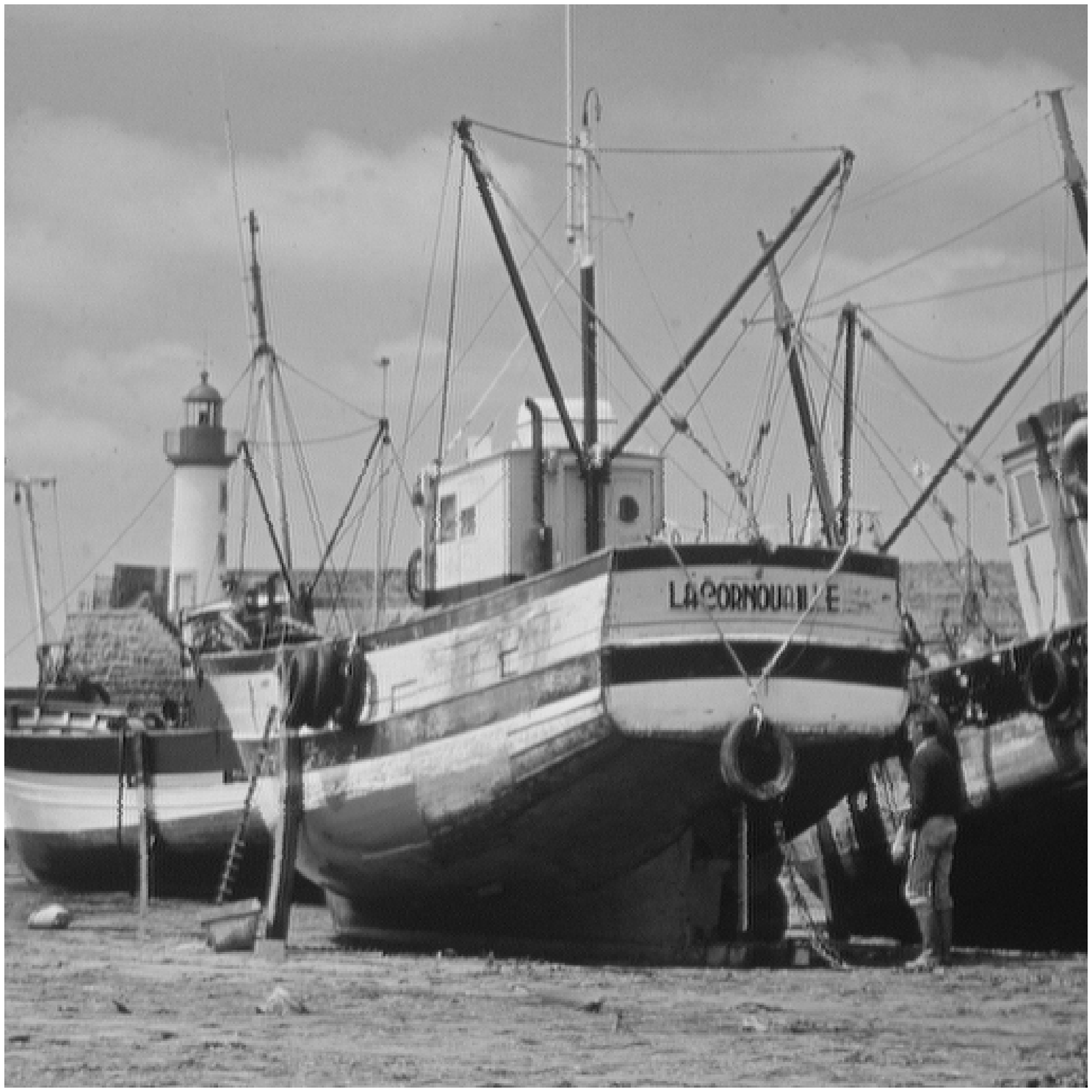}\phantom{2} \includegraphics{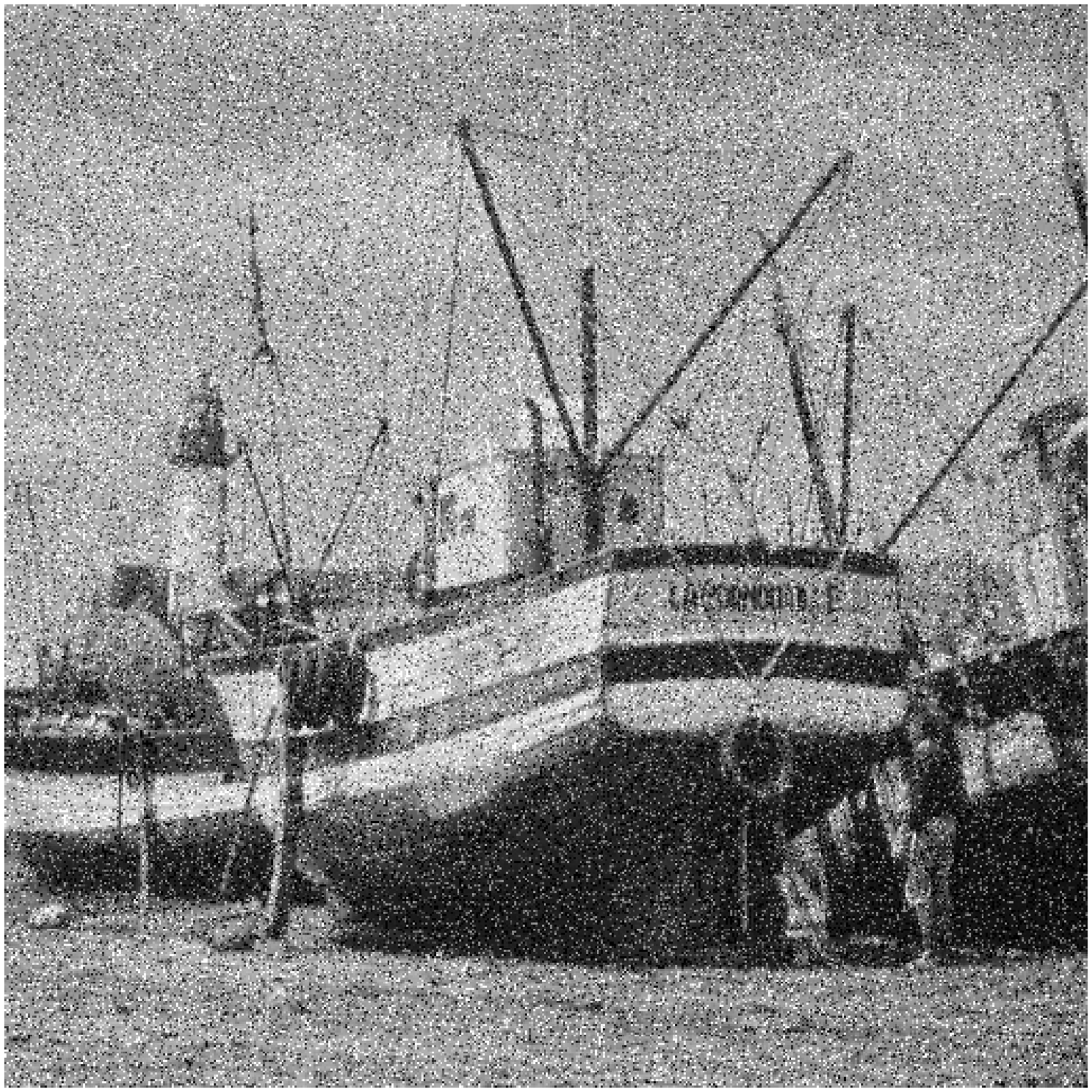}\phantom{2}
\includegraphics{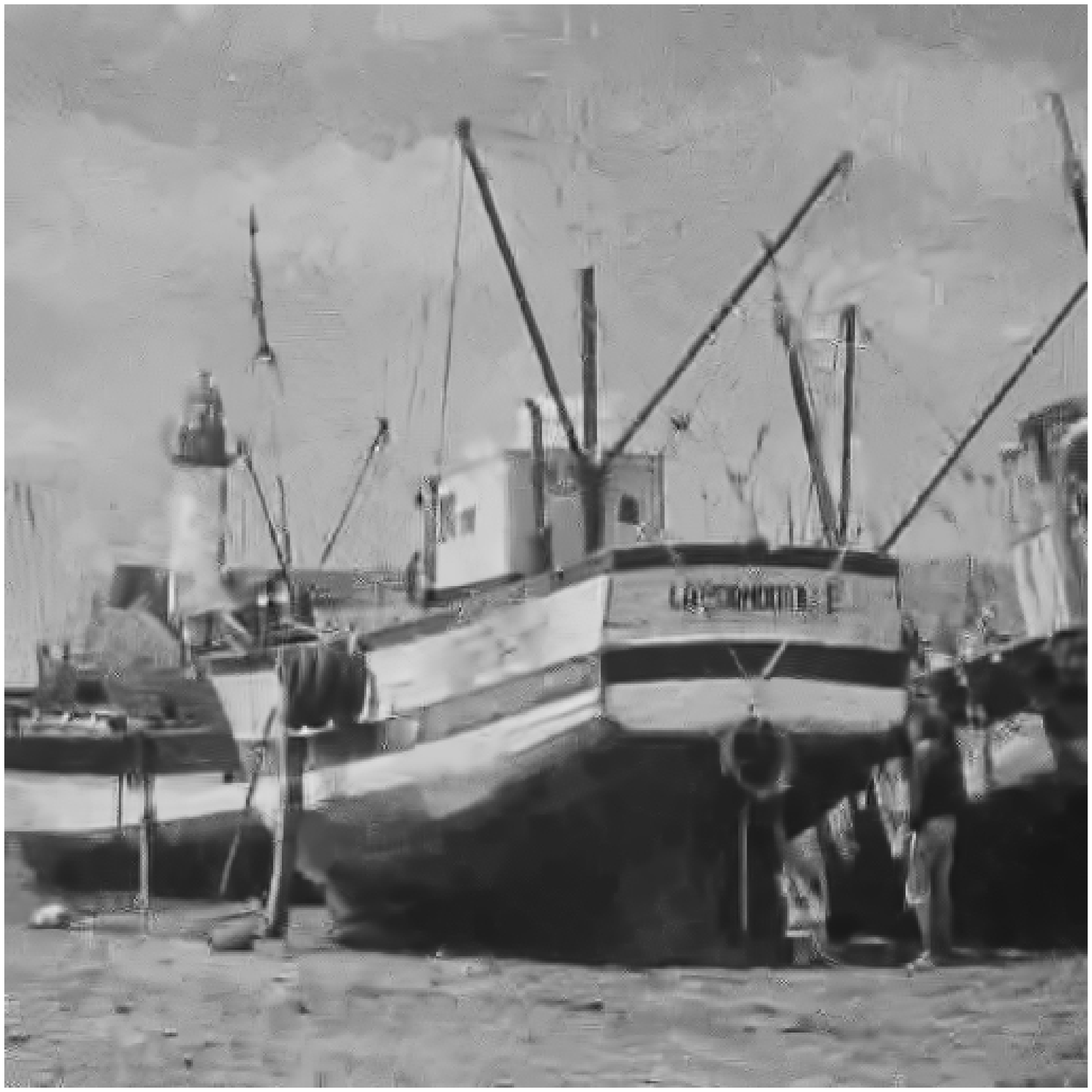}\phantom{2}
\includegraphics{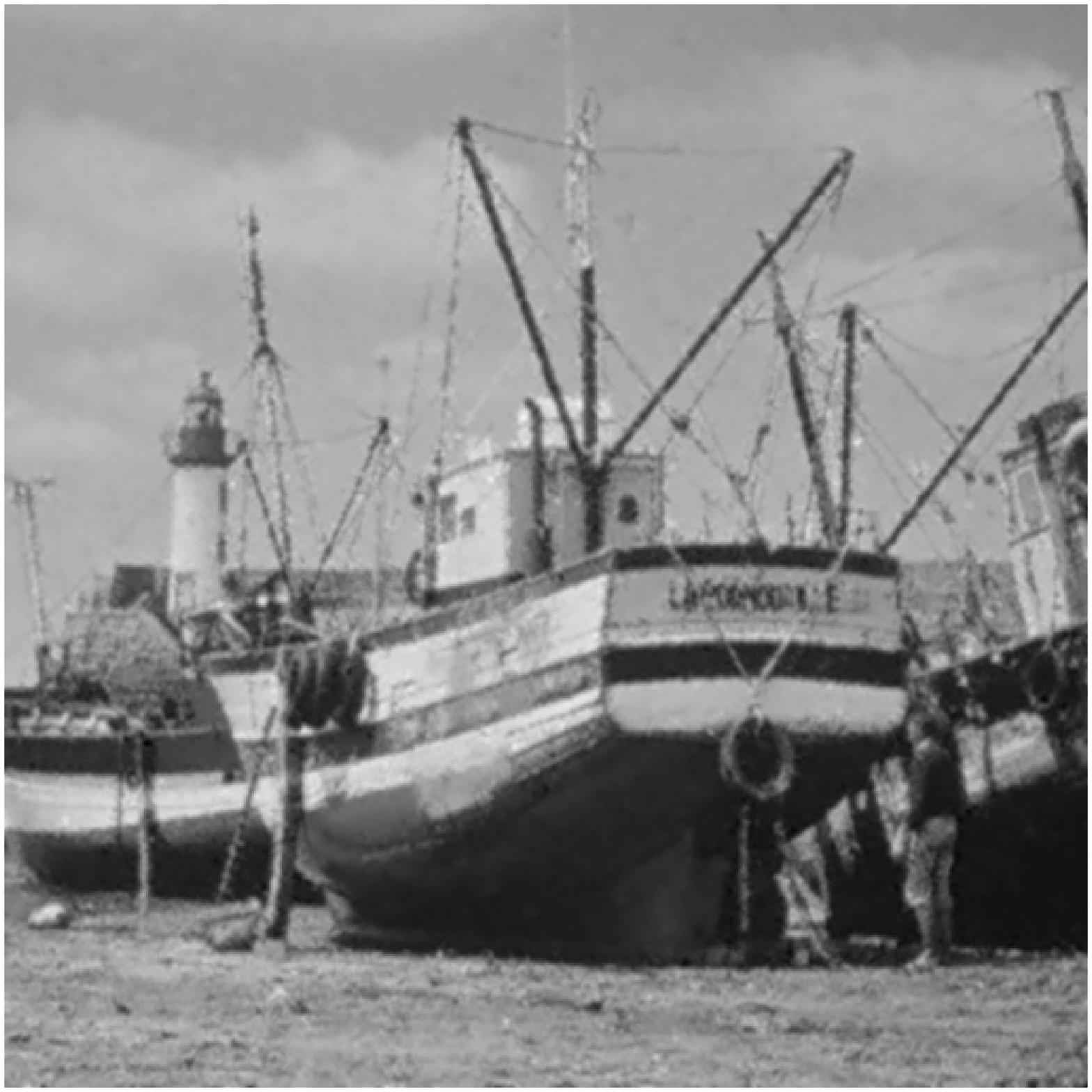}}
\centerline{(e) \hspace{13em}(f) \hspace{13em} (g) \hspace{13em} (h)}

\caption{(a) Original Image, (b) Original with additive Gaussian Noise - PSNR=22.14 dB, (c) Wavelet BiShrink Denoising \cite{Poly} - PSNR=31.17 dB, (d) Kernelized Denoising - PSNR=31.12 dB,
(e) Original Image, (f) Original with additive Impulse Noise - PSNR=15.52 dB, (g) BM3D Denoising \cite{DaFoKatEg} - PSNR=27.26 dB, (h) Kernelised Denoising - PSNR=29.14 dB.}\label{FIG:lena}
\end{sidewaysfigure*}

Figure \ref{FIG:lena} and tables \ref{TAB:impulse}, \ref{TAB:gaussian} show the obtained results using our algorithm on the Lena and Boat ($512\times 512$) grayscale images. More experimental results, the code in C (for the proposed methodology), as well as details on the implementation may be found at \url{http://cgi.di.uoa.gr/~stheodor/ker_den/index.htm}.  The results were compared with those obtained using several state of the art wavelet-based denoising packages, which are available on the internet (\cite{DaFoKatEg, TaFaMil, ElAh06, BuaCoMor, Poly}). The experiments show that the kernel approach performs equally well  as the well-known BiShrink wavelet-based method \cite{Poly} in the presence of Gaussian noise. However, it outperforms significantly the other denoising methods when impulse noise or mixed noise are considered (see figure \ref{FIG:lena}).  This enhanced performance is obtained at the cost of higher complexity, which is basically contributed by the optimization step, which is of the order of $O(MN)$ per pixel. Currently, more efficient optimization algorithms are considered. Moreover, the whole setting is open to a straightforward parallelization, when a parallel processing environment is available. This is also currently under consideration.

\section{Conclusions}
A novel denoising algorithm was presented based on the theory of RKHS. The semiparametric Representer Theorem was exploited in order to cope with the problems associated with the smoothing around edges, which is a common problem in almost all denoising algorithms. The comparative study against other denoising techniques, showed that significantly enhanced results are obtained in the case of impulse noise and mixed noise.

\bibliographystyle{latex8}
\bibliography{refs}
\end{document}